# Optimal Point-to-Point Trajectory Tracking of Redundant Manipulators using Generalized Pattern Search

**Atef A. Ata & Thi Rein Myo**
Mechatronics Engineering Department, Faculty of Engineering, International Islamic University Malaysia
Kuala Lumpur, Malaysia, atef@iiu.edu.my

*Abstract:* Optimal point-to-point trajectory planning for planar redundant manipulator is considered in this study. The main objective is to minimize the sum of the position error of the end-effector at each intermediate point along the trajectory so that the end-effector can track the prescribed trajectory accurately. An algorithm combining Genetic Algorithm and Pattern Search as a Generalized Pattern Search GPS is introduced to design the optimal trajectory. To verify the proposed algorithm, simulations for a 3-D-O-F planar manipulator with different end-effector trajectories have been carried out. A comparison between the Genetic Algorithm and the Generalized Pattern Search shows that The GPS gives excellent tracking performance.
*Keywords:* optimal trajectory, redundant, pattern search, genetic algorithms.

## 1. Introduction

The problem of designing optimal trajectory for redundant manipulators has attracted many researchers for the last three decades. One of the main reasons is the use of kinematically redundant robots is expected to increase in the future due to their increased flexibility. Some of the extra capabilities include the ability to avoid internal singularities or external obstacles over their entire workspace (Parket *et al.*,1989). Also, the inverse kinematics problem is underdetermined and admits an infinite number of distinct feasible solutions, meaning that a given end-effector poses can be realized by an infinite number of distinct manipulator configurations (McAvoy, *et al*, 2000). In order to overcome the shortcomings inherent in non-redundant robots, redundant robots have been utilized in industrial applications to increase flexibility and dexterity around a restricted task space in presence of obstacle. They can also provide a better ability to avoid singular configuration and the excessive velocities and accelerations encountered at singularities (Tian, L. & Collins, C., 2003).

Generally, there are three main approaches for trajectory planning for redundant manipulators, pseudo-inverse of jacobian matrix, variational approach, and optimization techniques based on the direct kinematics.

Whitney ( Whitney, E., 1969) introduced the pseudo-inverse approach, showed that the pseudo-inverse solution results in joint velocities having a minimum Euclidean norm. Some other researchers believe that this approach has many drawbacks. Klein and Huang (Klein, C. &Huang, C., 1983) showed that the pseudo-inverse is non-integrable, leading to manipulator joint space drift during cyclical tasks. Duffy (Duffy, J., 1990) showed that the pseudo inverse gives meaningless results in the case of a manipulator with different joint types. In addition to that the algorithm must take into account the problem of kinematic singularities that may be hard to tackle (Solteiro Pires *et al.*,2001)

Hirakawa and Kawamura (Hirakawa, A. & Kawamura, A., 1990) proposed a variational approach and B-spline curve for minimization of the consumed electrical energy to generate trajectory for redundant manipulators. The application of this method is oriented to repeated jobs by industrial robot manipulators with diminishing of long calculation time and difficulty to design the minimization toward vector.

To avoid all these drawbacks, Genetic Algorithms approaches (GAs) were introduced by Goldberg (Goldberg, E., 1989). A genetic algorithm is a stochastic search algorithm which can optimize nonlinear functions using the mechanics of natural genetics and natural selection. Several researchers have been carried out implementations of GAs in the field of robot trajectory planning. Parker et. al. (1989) presented a genetic algorithm approach which allows additional constraint to be specified easily. This approach was applied to test problems in which the maximum joint displacement in a point-to point positioning task is minimized. Davidor (Davidor, Y., 1991) applied a GA to generate the robot trajectory by finding the inverse kinematics for predefined end-effector robot paths. A trajectory of a 3-link planar redundant robot is simulated by minimizing the sum of the position errors at each of the knot points along the path. Yun and Xi (Yun, W-M & Xi, Y-G., 1990) presented a new method for optimum motion planning based on an improved genetic algorithm. This approach incorporates kinematics constraints, dynamics constraints as well as control constraints. Simulation results for two and three degree-of–freedom robots were presented. Hirakawa and Kawamura (Hirakawa, A. & Kawamura, A., 1996) proposed a combination of B-spline trajectory generation and steepest gradient optimization to design an optimal motion planning for redundant manipulators. However, the proposed optimization approach needs to determine the gradients of the objective function. McAvoy et al. (2000) proposed an approach utilizing genetic algorithms for optimal point-to-point motion planning for kinematically redundant manipulators to satisfy both the initial conditions and some other specified criteria. Their approach combines B-spline curves for the generation of smooth trajectories with genetic algorithms for optimal solution. Tian and Collins (2003) proposed a genetic algorithm using a floating point representation to search for optimal end-effector trajectory for a redundant manipulator. An evaluation function based on multiple criteria such as total displacement of all joints and the uniformity of Cartesian and joint space velocities was introduced. To verify their approach, simulations are carried out in free space and in a workspace with obstacles.

## 2. Problem Formulation

Consider the three-link planar manipulator shown in Figure (1) which has one extra degree of freedom to perform the operation.

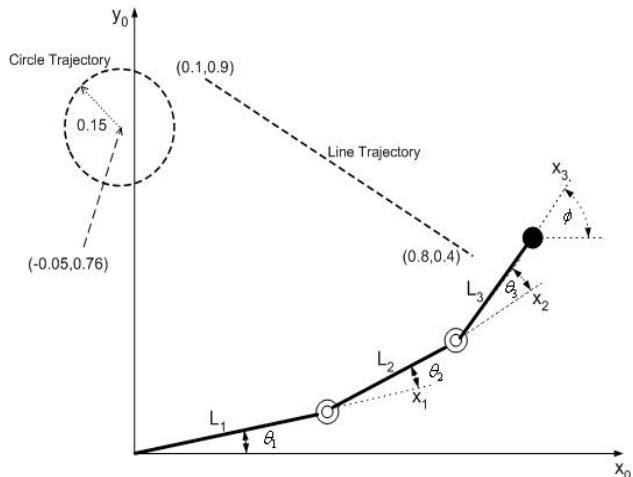

Fig.1. Three-link planar robot configuration

The joint angles $\theta_1, \theta_2$ and $\theta_3$ are assigned based on Denavit-Hartenberg representation and $\phi$ is the angle the end-effector makes with $X_0$ ($\phi = \theta_1 + \theta_2 + \theta_3$). $L_1$, $L_2$ and $L_3$ are the lengths of respected links. The end-effector is required to follow two trajectories accurately. The first one is straight line starting from (0.8, 0.4) and ending at (0.1, 0.9) while the second trajectory is a circle centered at (-0.05, 0.76) with a radius of 0.15 m.

## 3. Analytical Solution (10pt, bold)

Inverse kinematics is the analysis or procedure used to compute the joint coordinates for a given set of end effector coordinates. Basically, this procedure involves solving a set of equations which are, in general, nonlinear and complex. Although it is possible to solve the nonlinear equations, uniqueness is not guaranteed. Here, the analytical solution for inverse kinematics will be derived as a reference for comparisons with the optimal solutions. The three kinematics equations can be written as,

$$x = l_1 \cos\theta_1 + l_2 \cos(\theta_1 + \theta_2) + l_3 \cos(\theta_1 + \theta_2 + \theta_3) \quad (1)$$

$$y = l_1 \sin\theta_1 + l_2 \sin(\theta_1 + \theta_2) + l_3 \sin(\theta_1 + \theta_2 + \theta_3) \quad (2)$$

$$\phi = \theta_1 + \theta_2 + \theta_3 \quad (3)$$

Since the inverse kinematics problem is underdetermined, an infinite number of solutions exist depending upon the value angle $\phi$. So the angle $\phi$ can be assumed as a cubic polynomial in the form:

$$\phi = C_0 + C_1 t + C_2 t^2 + C_3 t^3 \quad (4)$$

in which the angular velocity is zero at the beginning and at the end of the task (rest-to-rest manoeuvring). This trajectory starts from initial value of 30° and ends at 70° within 5 seconds. The length for three links are $L_1 = 0.4$ m, $L_2 = 0.3$ m and $L_3 = 0.3$ m. On the other hand the trajectory of the end-effector can be divided into 20 via points so the x and y coordinates for each point are known. The analytical expressions for the joint angles $\theta_1$, $\theta_2$ and $\theta_3$ in terms of the Cartesian coordinates can be found as follows,

$$\theta_2 = a\tan 2(\sin\theta_2, \cos\theta_2) \quad (5)$$

Where, $\cos\theta_2 = \dfrac{x_n^2 + y_n^2 - l_1^2 - l_2^2}{2l_1 l_2}$,

$\sin\theta_2 = \pm\sqrt{1 - \left(\dfrac{x_n^2 + y_n^2 - l_1^2 - l_2^2}{2l_1 l_2}\right)^2}$

$$\theta_1 = a\tan 2\left(\frac{y_n(l_1 + l_2 \cos\theta_2) - x_n l_2 \sin\theta_2,}{x_n(l_1 + l_2 \cos\theta_2) + y_n l_2 \sin\theta_2}\right) \quad (6)$$

$$\theta_3 = \phi - (\theta_1 + \theta_2) \quad (7)$$

The simulation is performed using analytical method for both line and circle trajectories only line trajectory is presented in Figures 2a and 2b for comparison.

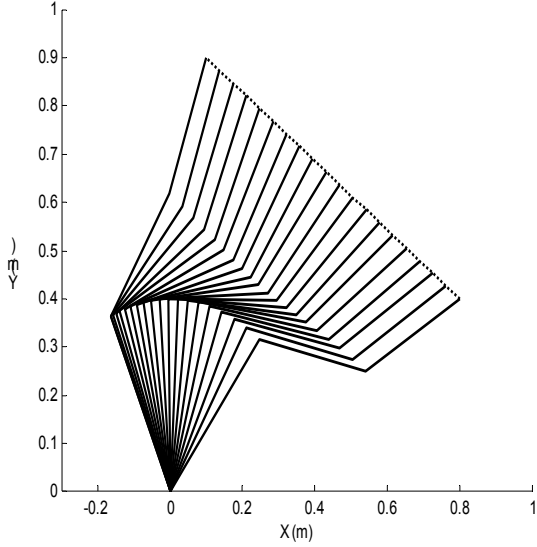

Fig. 2a. Robot configuration using analytic method

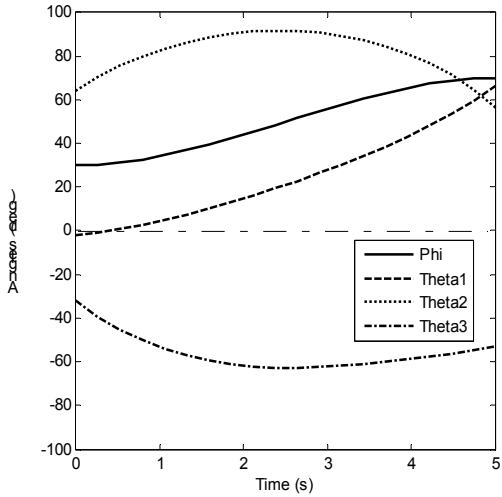

Fig. 2b. Joint angles using analytic method

**2. Simulation of Optimal Trajectory**

4.1 *Genetic Algorithm*
The real-coded in double precision genetic algorithm is used in this paper. The evaluation or fitness function is defined based on end-effector positioning error and joint angle displacements from the previous position satisfying Cartesian and joint velocity uniformity. The fitness function is defined as follows:

$$F_{fit} = C_1 E_e + C_2 D_j + C_3 V_e + C_4 V_j \quad (8)$$

Where, $E_e$ is the error between desired position and generated position of end effector. $D_j$ is the joint displacements between successive points. $V_e$ and $V_j$ are velocities of end-effector and joints of robot manipulator. $C_1$, $C_2$, $C_3$ and $C_4$ are weighting factors to control the desired configuration which satisfy the constraint $C_1 + C_2 + C_3 + C_4 = 1$. Since the objective is to minimize the error between the desired and generated position of end-effector, $E_e$ will be defined as

$$E_e = \sum_{i=1}^{n} \sqrt{(x^i - x_g^i)^2 + (y^i - y_g^i)^2} \quad (9)$$

Where, $(x^i, y^i)$ are desired end-effector positions and $(x_g^i, y_g^i)$ are generated end-effector positions.

The joint displacements between successive points are considered in evaluation function in order to minimize actuator motions. To minimize the joint movements along the trajectory, the function will be

$$D_j = \sum_{i=1,k=3}^{i=n,k=3} \left(\theta_k^{i+1} - \theta_k^i\right)^2 \quad (10)$$

Since the trajectory is divided into equal lengths between successive points, the velocities for end-effector and joint displacements will be

$$V_e = \overrightarrow{P_i P_{i+1}} \to const \quad (11)$$

$$V_j = \sum_{i=1,k=3}^{i=n,k=3} \left|\theta_k^{i+1} - \theta_k^i\right| \to const \quad (12)$$

Weighting factors are defined as $C_1= 0.4$, $C_2= 0.1$, $C_3= 0.3$, $C_4 = 0.2$ for line trajectory and $C_1 = 0.7$, $C_2= 0.1$, $C_3= 0.1$, $C_4 = 0.1$ for circle trajectory. The initial joint space configurations are assumed as $\theta_1^I = 60°, \theta_2^I = -30°$ and $\theta_3^I = -30°$ for line trajectoryand for circle trajectory. The parameters for GA are listed in table 1.

| Population | 100 |
|---|---|
| Fitness scaling | Rank |
| Selection | Stochastic Uniform |
| Reproduction | Elite count = 2, crossover rate = 1.2 |
| Mutation | Gaussian(scale = 1, shrink =1) |
| crossover | Scattered function |
| Migration | Forward (fraction 0.2, Interval =20) |
| Generation | 200 |

Table 1. Genetic Algorithm parameters

The simulations are carried out for robot configurations, angles profiles and error for both trajectories and the results are presented at Figures 3 and 4 for line and circle end-effector trajectories respectively. All simulations have been done using MATLAB Genetic Algorithm and Direct Search Toolbox.

### 4.2 *Genetic Algorithm and Pattern Search (Generalized Pattern Search)*

Pattern search method is a class of direct search method to solve for nonlinear optimization. A pattern search algorithm computes a sequence of points that get closer and closer to the optimal point. At each step, the algorithm searches a set of points, called a *mesh*, around the *current point* — the point computed at the previous step of the algorithm. The algorithm forms the mesh by adding the current point to a scalar multiple of a fixed set of vectors called a *pattern*. If the algorithm finds a point in the mesh that improves the objective function at the current point, the new point becomes the current point at the next step of the algorithm [13].

Optimal points from GA are introduced into Pattern Search Algorithm as inputs to get better result. Pattern Search is only used for reducing the error of end-effector positions in trajectory, the evaluation function is defined as in equation (9) as.

$$F_{eval} = \sum_{i=1}^{n} \sqrt{(x^i - x_g^i)^2 + (y^i - y_g^i)^2} \qquad (13)$$

Where, $(x^i, y^i)$ are desired end-effector positions and $(x_g^i, y_g^i)$ are generated end-effector positions from Pattern Search tool. The complete poll, consecutive polling order, and complete search methods are used in Pattern Search Algorithm. The simulations are carried out for robot configurations, angles profiles and tracking error of end-effector for both trajectories. Figures 4 and 6 show the simulation results for line and circle end-effector trajectories respectively.

### 5. Discussion

It can be observed clearly from Figures 3, 4, 5 and 6 that adding Pattern Search to the Genetic Algorithm improves the final results and reduces the total tracking error considerable. Comparing with the analytical solution, GPS algorithm gives almost the same results with zero tracking error. This improvement makes it possible for the end-effector to track the desired trajectory accurately. In case of constrained motion where the force exerted by the end-effector depends mainly on the distance between the end-effector and the constraint surface, this algorithm minimizes the contact force to a great extent. In terms of optimized joint angles, there is no considerable differenc between the two algorithms. The genetic algorithm searches for the global minimum and the patern search refines the local minimum reached by the genetic algorithm. Our conclusions agree with the recent results indicating that adding Coordinate Search Algorithm to the simple Genetic Algorithm to enable clear convergence and avoid the risk of getting attracted by local minimum. This combination constitutes a Generalized Pattern Search that uses the sGA for the global search and uses the Coordinate Search for the local search (The Math Works, 2004).

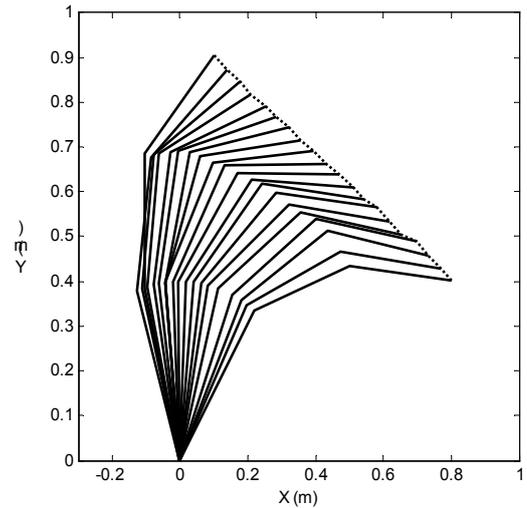

Fig. 3a. Robot configuration

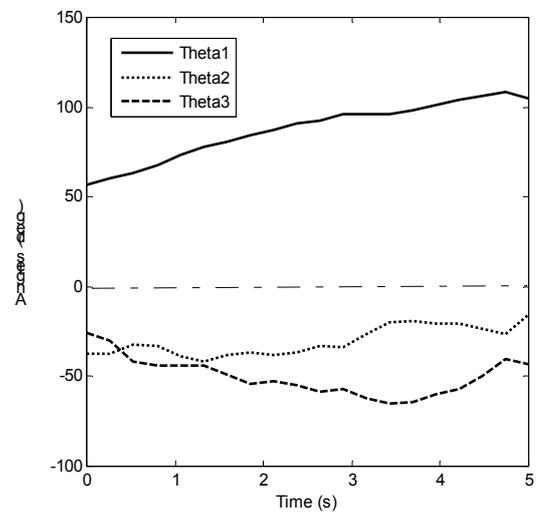

Fig. 3b. Optimized angles

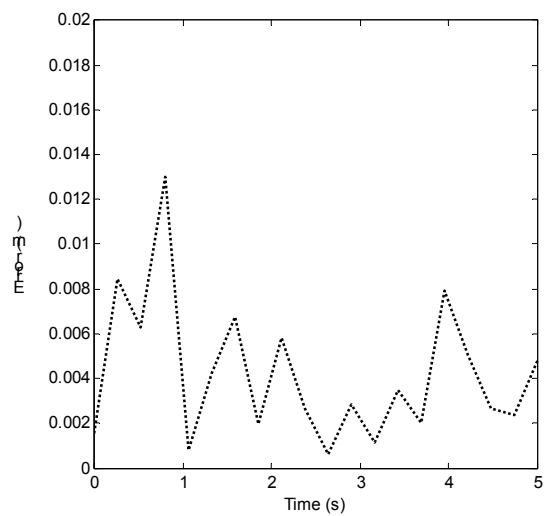

Fig. 3c. Tracking error

Figs. 3. Line trajcetory using Genetic Algorithm

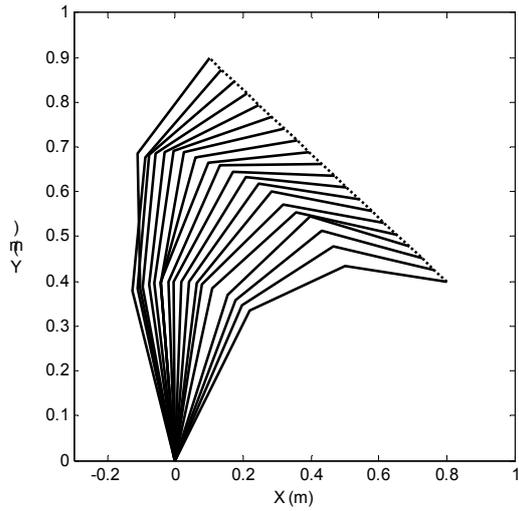

Fig. 4a. Robot configuration

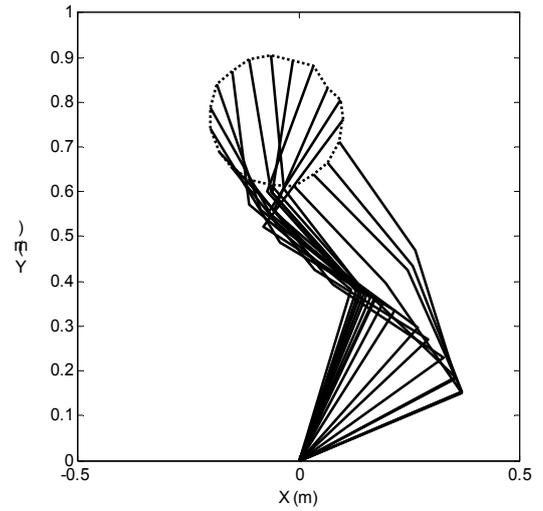

Fig. 5a. Robot configuration

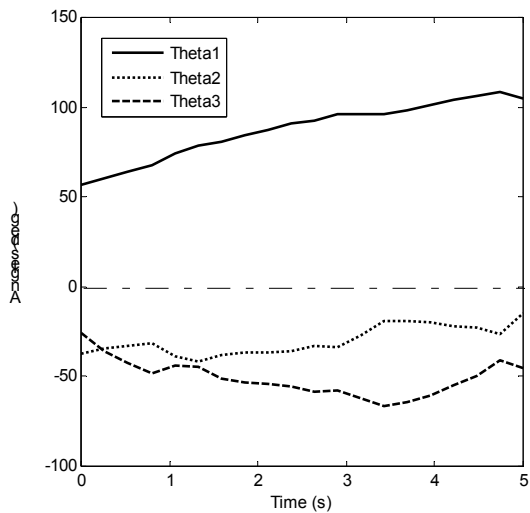

Fig. 4b. Optimized angles

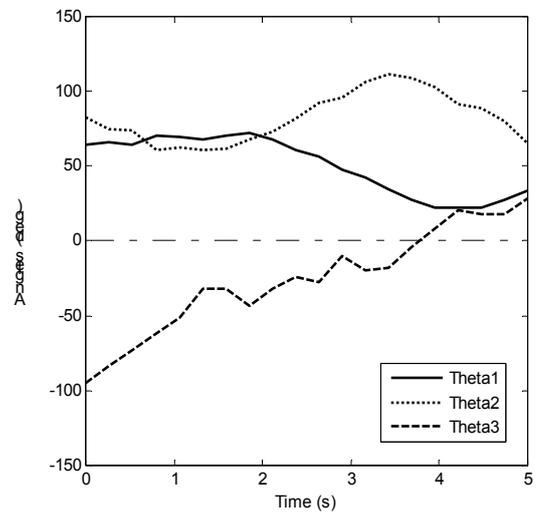

Fig. 5b. Optimized angles

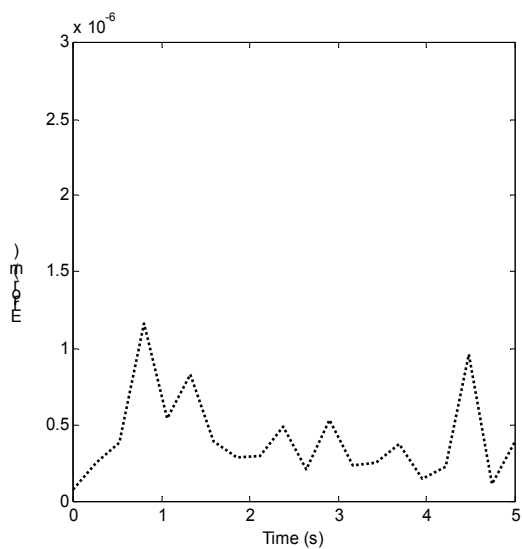

Fig. 4c. Tracking error

Figs. 4. Line trajcetory using Genetic Algorithm & Pattern Search

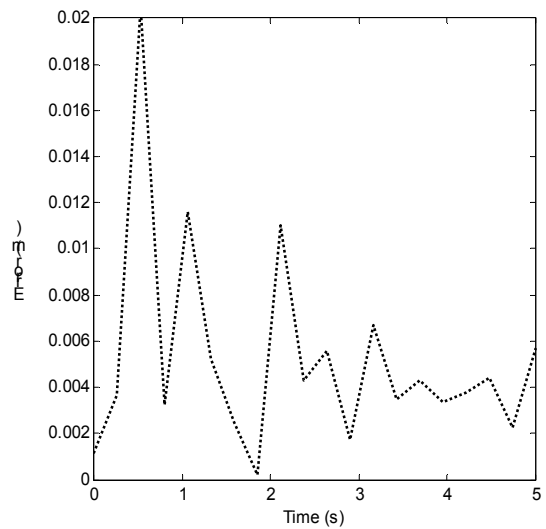

Fig. 5c Tracking error

Figs. 5. Circle trajcetory using Genetic Algorithm

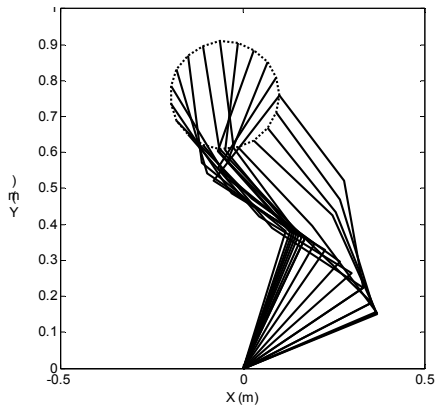

Fig. 6a Robot configuration

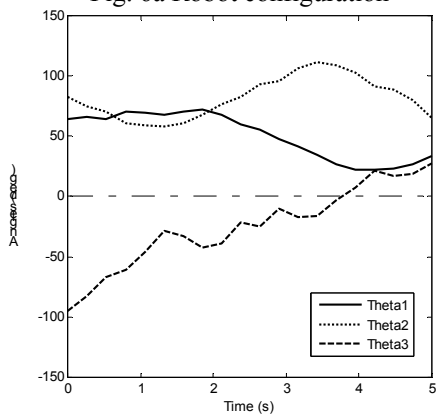

Fig. 6b Optimized angles

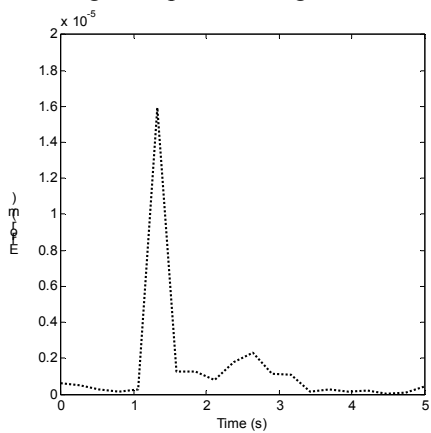

Fig. 6c Tracking error

Figs. 6 Circle trajcetory using Genetic Algorithm & Pattern Search

## 6. Conclusions

A new algorithm is introduced for the optimal trajectory tracking of planar redundant manipulators. This algorithm which combines Genetic Algorithm and Pattern Search can be considered as a Generalized Pattern Search algorithm. It has been applied successfully for two end-effector trajectories with excellent tracking performance.

**Acknowledgement:** The authors would like to thank the Research Center , International Islamic University Malaysia for supporting this research